# A Review of Graph Neural Networks and Their Applications in Power Systems

Wenlong Liao, Birgitte Bak-Jensen, Jayakrishnan Radhakrishna Pillai, Yuelong Wang, and Yusen Wang

*Abstract*—Deep neural networks have revolutionized many machine learning tasks in power systems, ranging from pattern recognition to signal processing. The data in these tasks is typically represented in Euclidean domains. Nevertheless, there is an increasing number of applications in power systems, where data are collected from non-Euclidean domains and represented as graph-structured data with high dimensional features and interdependency among nodes. The complexity of graph-structured data has brought significant challenges to the existing deep neural networks defined in Euclidean domains. Recently, many publications generalizing deep neural networks for graph-structured data in power systems have emerged. In this paper, a comprehensive overview of graph neural networks (GNNs) in power systems is proposed. Specifically, several classical paradigms of GNNs structures (e.g., graph convolutional networks) are summarized, and key applications in power systems, such as fault scenario application, time series prediction, power flow calculation, and data generation are reviewed in detail. Furthermore, main issues and some research trends about the applications of GNNs in power systems are discussed.

## Nomenclature

*Abbreviations*

| | |
|---|---|
| RES | renewable energy sources |
| PV | photovoltaic |
| DNNs | deep neural networks |
| RNNs | recurrent neural networks |
| CNNs | convolutional neural networks |
| DBNs | deep belief networks |
| GANs | generative adversarial networks |
| AEs | automatic encoders |
| GNNs | graph neural networks |
| GCNs | graph convolutional networks |
| GRNNs | graph recurrent neural networks |
| GATs | graph attention networks |
| GGNs | graph generative networks |
| STGNNs | spatial-temporal graph neural networks |
| GRL | graph reinforcement learning |
| GTL | graph transfer learning |
| RL | reinforcement learning |
| ChebNet | Chebyshev spectral CNN |
| DCNN | diffusion-convolutional neural network |
| DGCN | dual-graph convolutional network |
| LSTMs | long short-term memories |
| GRUs | gated recurrent units |
| ReLU | rectified linear unit |
| LeakyReLU | leaky rectified linear unit |
| GaAN | gated attention network |
| GAEs | graph automatic encoders |
| VGAEs | variational graph auto-encoders |
| GCRNs | graph convolutional recurrent networks |
| MLPs | multi-layer perceptions |
| SVMs | support vector machines |
| KNNs | k-nearest neighbors |
| PCAs | principal component analyses |
| RFs | random forests |
| TCN | temporal convolutional network |
| RMSE | root mean squared error |
| MAE | mean absolute error |
| MAPE | mean absolute percentage error |

*Parameters*

| | |
|---|---|
| $G=(V, E)$ | the graph-structured data |
| $V$ | a set of nodes |
| $E$ | a set of edges |
| $N(v_i)$ | the neighborhood of the $i^{th}$ node |
| $X^{node}$ | a nodal feature matrix |
| $X^{edge}$ | a feature matrix of edges |
| $A$ | an adjacency matrix |
| $\tilde{A}$ | a reconstructed adjacency matrix |
| $g \in R^n$ | a filter |
| $x \in R^n$ | a sample |
| $\sigma_{FP}, \sigma_{DC}, \ldots$ | the activation functions for different models |
| $g_W$ | the filter parameterized by $W$ |
| $I_n$ | identity matrix of dimension $n$ |
| $\Lambda$ | a diagonal matrix |
| $U \in R^{n \times n}$ | the eigenvectors ordered by eigenvalues |
| $W' \in R^K$ | a vector that consists of Chebyshev coefficients |
| $\mathscr{F}(\cdot)$ | the Fourier transform |
| $\mathscr{F}^{-1}(\cdot)$ | the inverse Fourier transform |
| $L$ | the Laplacian matrix |
| $\hat{D}$ | a diagonal matrix of node degrees |
| $W^*$ | the filter parameters |
| $W^i, U^i, W^o, U^o \ldots$ | the weights of different gates |
| $b^i, b^o, \ldots$ | the bias vectors of different gates |
| $\theta_D, \theta_G, \ldots$ | the parameters of different models to be learned |
| $H_{DC}^t, H_{DG}^t, \ldots$ | the hidden state of different models at time step $t$ |
| $t$ | the time step |
| $*$ | graph convolutional operation |
| $\odot$ | Hadamard product operation |
| $\|$ | concatenation operation |

## I. Introduction

### A. Background

After several decades of development, the smart grid has evolved into a typical dynamic, non-linear, and large-scale control system, known as the power system. The multi-directional information makes it hard to find optimal solutions that coordinate all participants, such as distribution

W. Liao, B. Jensen, and J. Pillai are with the Department of Energy Technology, Aalborg University, Aalborg, Denmark (e-mail: weli@et.aau.dk; bbj@et.aau.dk; jrp@et.aau.dk).
Y. Wang is with the State Grid Tianjin Chengxi Electric Power Supply Branch, Tianjin, China (e-mail: yuelong.wang@tj.sgcc.com.cn).
Y. Wang (corresponding author) is with the School of Electrical Engineering and Computer Science, KTH Royal Institute of Technology, Stockholm, Sweden (e-mail: yusenw@kth.se).

systems operators, producers, demand response aggregators, and consumers [1]. For example, the high penetration of renewable energy sources (RES), such as photovoltaic (PV) plants and wind farms, brings fluctuation and intermittence to power systems, which requires more reserve capacity to avoid power outage. The integration of flexible sources (e.g., electric vehicles) poses revolutionary changes to radial distribution networks, such as relay protection, bidirectional power flow, and voltage regulation [2]. Moreover, the deregulation of electricity markets makes it difficult to find a strategy that is beneficial to both customers and producers. In these cases, traditional model-based methods are hard to fully meet the control and analysis requirements of power systems because of their uncertainty and complexity. For example, traditional model-based methods for scenario generations of RES aren't able to accurately capture the probability distribution characteristics and fluctuations of power curves [3], since they need to artificially assume the probability density function of power curves. Furthermore, traditional model-based methods are not universal, since the probability distributions of power curves vary from regions and times.

The outstanding performances of deep neural networks (DNNs) in computer visions bring new opportunities to these problems that cannot be solved by traditional model-based methods in power systems. Many challenging tasks, such as time series prediction of loads and RES, fault diagnosis, scenario generation, and operational control, which is highly dependent on hand-made feature engineering to extract information-rich latent features, have recently been completely changed by a variety of DNNs, such as recurrent neural networks (RNNs), convolutional neural networks (CNNs), generative adversarial networks (GANs), and automatic encoders (AEs) [4]. The successful applications of DNNs in many tasks of power systems are due in part to the rapid development of advanced sensors, smart meters, computing resources, and the effectiveness of DNNs that mines potential representations from various Euclidean data, such as power curves of RES, dissolved gas of power transformers, and images of insulators. Taking the detection of power line insulator defects as an example, the image can be considered as a regular grid in Euclidean domains as shown in Fig 1(a). To accurately identify different states of insulators, convolutional filters of CNNs are used to extract locally latent features due to their advantages in processing compositionality, local connectivity, and shift-invariance of 2-D data [5]. However, there are some data of power systems recorded from non-Euclidean domains, such as the graph-structured data with nodes and edges. For instance, the input data of power flow calculation includes loads of each node and an adjacency matrix, which is a kind of graph-structured data as shown in Fig 1(b) [6]. The complex graph-structured data has posed huge challenges to existing DNNs, which are defined in Euclidean domains [7]. Specifically, the graph-structured data may have unordered nodes of different sizes, and the number of neighbors of these nodes may be different, since the graph-structured data may be irregular. Although some crucial operations (e.g., convolution) are easy to calculate in Euclidean domains, they are hard to be generalized to graph domains. In this case, the existing DNNs in Euclidean domains, such as CNNs and RNNs are not suitable for processing the graph-structured data [8], since they stack the features of nodes by a specific order and ignore the topological information.

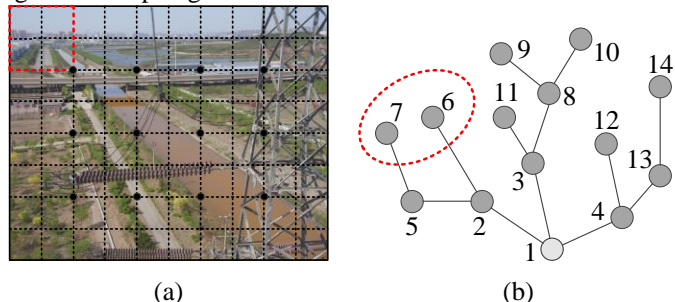

Fig. 1. Euclidean convolution versus graph convolution. (a) Detection of insulators defects in Euclidean domains. (b) Power flow calculation of the IEEE 14-bus system in graph domains.

Recently, the generalization of DNNs from Euclidean domains to graph domains has received more and more attention. Various new paradigms and definitions of DNNs in graph domains have been rapidly updated to deal with the complicated graph-structured data in the past few years. Fig. 2 visualizes the number of published papers about GNNs from 2010 to 2020 through the advanced search function of the Google Scholar, which evidently shows strong growth over the last three years. The classical graph neural networks (GNNs) mainly include graph convolutional networks (GCNs), graph recurrent neural networks (GRNNs), graph attention networks (GATs), graph generative networks (GGNs), spatial-temporal graph neural networks (STGNNs), and hybrid forms of GNNs, such as graph reinforcement learning (GRL) and graph transfer learning (GTL) [9], which have shown outstanding performance for the graph-structured data.

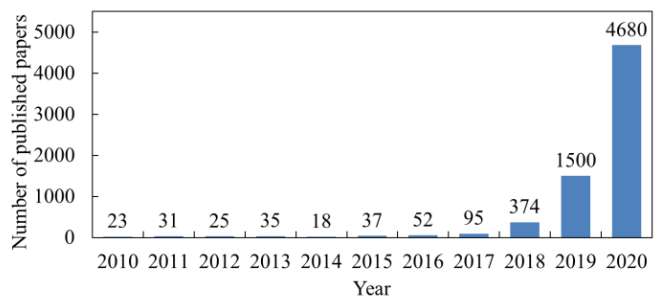

Fig. 2. The yearly number of published articles about GNNs.

There is a limited number of existing review articles related to GNNs or deep learning in power systems. These reviews either focus on the structures of GNNs and their applications in computer visions, or analyze the applications of traditional DNNs in power systems. For example, the training strategies and model architectures related to five different types of GNNs are discussed in [10]. A few review articles pay close attention to the application for non-structural scenarios, such as image classification [11] and natural language processing [12]. Some methods for how much various models can be trained on large-scale knowledge graphs are reviewed in [13]. The theoretical background of GNNs and some geometric data, such as social networks and point clouds are introduced in [14]. Reinforcement learning (RL) is becoming increasingly popular because of its success in dealing with challenging decision-making problems in power systems. The recent combinations of RL and DNNs in Euclidean domains, and their application in power systems are critically reviewed in [15], [16]. A comprehensive review of the advantages of deep representation learning is conducted in [17], which covered several large ranges, including supervised, unsupervised, and semi-supervised applications. In general, a part of the existing articles only focus on the application of GNNs in computer science (e.g., Recommendation systems, link prediction, and protein structure classification), but do not review the applications in power systems. Another part of the articles investigate the advantages of traditional DNNs in power systems, but does not involve GNNs. Relatively, this paper provides a comprehensive review of classical GNNs and their applications in power systems for interested researchers who major in electrical engineering, machine learning, and energy.

*B. Brief History from Traditional DNNs to GNNs*

Although GNNs are still in their infancy, the traditional DNNs behind them have a very long history in Euclidean domains. The rapid development of traditional DNNs has greatly promoted the research process of GNNs, since many important operations in graph domains draw lessons from operations in Euclidean domains. The concept of GNNs was first proposed in 2005 [18] and further clarified in Ref. [19]. These early articles propagate neighboring information iteratively through the recurrent neural architecture to learn the representation of target nodes until stable fixed points are reached. This process involves huge a computational resource, and there have been many recent works focusing on this problem recently.

Motivated by the outstanding performance of CNNs in computer visions, the convolutional operation for graph-structured data was defined by many publications. In 2014, the first important research on spectral-based GCNs was proposed in [20], which used the spectral graph theory to develop a new variant of graph convolutional operation. From then on, various spectral-based GCNs have been continuously proposed and improved. The spatial theory is an important theoretical basis of another popular definition of graph convolution. The graph mutual dependence was solved by architecturally complex non-recursive layers presented in [21], while this important work was ignored at that time. Recently, spatial-based GCNs have developed rapidly, since spectral-based GCNs usually deal with the whole graphs simultaneously, and they are difficult to extend to large-scale graphs. In addition to GCNs, many other GNNs generalized from traditional DNNs have been developed over recent years. These models mainly include GRNNs, GATs, GGNs, STGNN, and hybrid forms of GNNs whose structures and parameters are given in Section II.

*C. Contributions and Organization*

This paper focuses on providing a comprehensive review of the GNNs and their applications in power systems, and identifies future challenges. The contributions of this paper are summarized as follows.

1) Introducing several classical paradigms of GNNs structures. Each paradigm provides detailed structure and descriptions of representative applications, and interested researchers can easily apply it to different fields.

2) Conducting a comprehensive survey of GNNs on power systems with the newest developments (e.g., fault scenario application, time series prediction, power flow calculation, data generation, and so forth), particularly over the past three years. Some possible extension works about the application of GNNs in power system is proposed.

3) Discussing the limitations of existing models, the theoretical advantage of GNNs, and possible future research directions in power systems.

The remainder of this paper is organized as follows. Section II introduces the definitions of graph-structured data and several classical paradigms of GNNs structures. Section III presents a comprehensive survey of GNNs on power systems with the newest developments. Section IV shows the current challenges and proposes application directions. Section V summarizes conclusions.

## II. DEFINITION AND PARADIGMS

*A. Definitions of Graph-Structured Data*

Normally, the graph-structured data is represented as $G=(V,E)$, where $E$ is a set of edges and $V$ is a set of nodes [22]. Specifically, $v_i$ is the $i^{th}$ node and $e_{ij}$ is the edge from the $j^{th}$ node to the $i^{th}$ node. For the $i^{th}$ node, its neighborhood can be denoted as $N(v_i)=\{u_i \in V|(v_i,u_i)\in E\}$. The graph-structured data generally has a nodal feature matrix $X^{node}$ of $n \times f$ scales, and a feature matrix $X^{edge}$ of $m \times c$ scales for edges. The adjacency matrix $A$ is a matrix of $n \times n$ scales where $a_{ij}$ is equal to 0 if $e_{ij} \notin E$ and $a_{ij}$ is equal to 1 if $e_{ij} \in E$.

For the spatial-temporal graph, it is an attributed graph where the features of nodes change with times [23]. A spatial-temporal graph can be represented as $G^{(t)}=(V,E,X^{(t)})$. For the directed graph, it has an asymmetric adjacency matrix, since these edges are directed from one node to another. Relatively, edges of the undirected graph are all undirected, i.e. the adjacency matrix is symmetric, and its normalized Laplacian matrix $L$ can be defined as:

$$L = I_n - D^{-\frac{1}{2}} A D^{-\frac{1}{2}} \quad (1)$$

where $D$ denotes the diagonal matrix of node degrees with $D_{ii} = \sum_{j=1}^{n} A_{ij}$; and $I_n$ is an identity matrix.

Since the normalized graph Laplacian matrix $L$ is real symmetric positive semi-definite, it can be factored as:

$$L = U \Lambda U^T \quad (2)$$

where $U \in R^{n \times n}$ denotes the corresponding eigenvectors ordered by eigenvalues $\lambda$; and $\Lambda$ denotes the diagonal matrix with $\Lambda_{ii} = \lambda_i$.

Before going further into the next sections, Table I lists the several popular paradigms of GNNs.

TABLE I
MAIN CHARACTERISTICS OF VARIOUS PARADIGMS

| Paradigms | Variants | Main function | Ref |
|---|---|---|---|
| GCNs | Spectral-based Nets | Extracting latent representation of the graph-structured data | [20],[24]-[27] |
| | Spatial-based Nets | | [29]-[32] |
| GRNNs | Graph GRUs | Solving the long-term dependencies of the graph-structured data | [39] |
| | Tree LSTM | | [40] |
| GATs | GAT | Incorporating attention mechanisms to the propagation step of GNNs | [43] |
| | GaAN | | [44] |
| GGNs | GAEs | Generating new graph-structured data | [46] |
| | GVAEs | | [47] |
| | GGANs | | [48] |
| STGNNs | RNN-based Nets | Learning hidden patterns from spatial-temporal graphs | [49]-[53] |
| | CNN-based Nets | | [54]-[56] |
| Hybrid forms of GNNs | GRL | Improving control and perception of graph-structured data | [59]-[61] |
| | GTL | Transferring of knowledge in graph domains | [62]-[64] |

*B. Graph Convolutional Networks*

This section will discuss GCNs that are generalized from Euclidean domains to graph domains. The existing GCNs mainly include two categories: spectral-based GCNs and spatial-based GCNs [24]. Since both categories of GCNs have a large number of variants, only a few classic models are listed to illustrate the principle and structure.

1) Spectral-based GCNs. Unlike the images in Euclidean domains, the graph-structured data does not have the characteristics of translation invariance, which makes it hard to directly define the convolutional operation in graph domains. In 2014, a spectral network was proposed in [20]. It transforms the samples into the Fourier domains to perform convolutional operations through Fourier transform, and then the samples are transformed back to the graph domains through inverse Fourier transform. Specifically, the graph convolutional operation of the sample $x \in R^n$ with a filter $g \in R^n$ can be defined as:

$$g * x = \mathcal{F}^{-1}(\mathcal{F}(x) \odot \mathcal{F}(g)) = U(U^T x \odot U^T g) = U g_W U^T x \quad (3)$$

where $*$ is the graph convolutional operation; $\odot$ is the Hadamard product operation; and $g_W = \text{diag}(U^T g)$ is the filter parameterized by $W \in R^n$.

In [25], the Chebyshev spectral CNN (ChebNet) that approximate $g_W$ by the truncated expansion of Chebyshev polynomials $T_k(x)$ up to $K^{th}$ order is proposed:

$$g * x \approx \sum_{k=0}^{K} W_k' T_k\left(\frac{2L}{\lambda_{max}} - I_n\right)x \quad (4)$$

$$T_k(x) = 2xT_{k-1}(x) - T_{k-2}(x), T_0(x) = 1, T_1(x) = x \quad (5)$$

where $W' \in R^K$ is a vector that consists of Chebyshev coefficients; and $\lambda_{max}$ is the largest eigenvalue. Because the operation is a $K^{th}$ order polynomial in the Laplacian, it is the K-localized.

Moreover, a new graph convolutional network is proposed to approximate ChebNet by assuming that $\lambda_{max} = 2$ and $K = 1$ [26]. Its mathematical formula is:

$$g * x \approx W_0' x - W_1' D^{-\frac{1}{2}} A D^{-\frac{1}{2}} x \quad (6)$$

To alleviate over-fitting problems and restrain the number of parameters, it further assumes that $W = W_0' = -W_1'$, which lead to the following formula:

$$g * x \approx W\left(I_n + D^{-\frac{1}{2}} A D^{-\frac{1}{2}}\right)x \quad (7)$$

Furthermore, the renormalization trick is utilized to avoid vanishing gradients problems in GCN [27]:

$$g * x \approx W \hat{D}^{-\frac{1}{2}} \hat{A} \hat{D}^{-\frac{1}{2}} X, \hat{A} = A + I_n \quad (8)$$

where $\hat{D}$ denotes the diagonal matrix of node degrees with $\hat{D}_{ii} = \sum_{j=1}^{n} \hat{A}_{ij}$.

Finally, the Eq. (8) can be generalized to the multi-channel convolution [11], [27]:

$$Z_{out} = \hat{D}^{-\frac{1}{2}} \hat{A} \hat{D}^{-\frac{1}{2}} XW'' \quad (9)$$

where $X \in R^{n \times c}$ denotes a input signal with $f$ filters and $c$ input channels; $Z_{out} \in R^{n \times f}$ denotes an output signal; and $W'' \in R^{c \times f}$ denotes the filter parameters.

2) Spatial-based GCNs. Analogous to the convolutional operation in the Euclidean domains, spatial-based GCNs directly define the convolutional operation on the graph domains by operating on spatially close neighbors. The key challenges of these spatial-based GCNs are to define convolutional operations with the different number of neighborhoods and to keep the local invariance [28], [29].

In 2015, Neural FPs defined the convolutional operation by using different weight matrices for nodes with different degrees [30]. Its mathematical formula is:

$$H_{FP,v}^{t} = \sigma_{FP}\left(X_{FP}^{t} W_t^{|N(v)|}\right), X_{FP}^{t} = H_{FP,v}^{t-1} + \sum_{i=1}^{|N(v)|} H_{FP,i}^{t-1} \quad (10)$$

where is $H_{FP,v}^{t}$ the hidden state of node $v$ of the Neural FPs at time step $t$; $\sigma_{FP}(\cdot)$ is the activation function of Neural FPs, such as rectified linear unit (ReLU); $N(v)$ is the neighborhood of node $v$; and $W_t^{|N(v)|}$ is a weight matrix with the degree $|N(v)|$ at time step $t$. The main shortcoming of this convolutional operation is that it cannot be applied to graphs of large-scale with large node degrees.

Moreover, a diffusion-convolutional neural network (DCNN) is proposed to define the neighborhood for nodes by transition matrices in [31]. Its mathematical formula is:

$$H_{DC}^{t} = \sigma_{DC}\left(W_{DC}^{t} \odot P_{DC} X_{DC}^{t}\right) \quad (11)$$

where $X_{DC}^{t} \in R^{n \times f}$ is the input data at time step $t$ ($f$ is the number of features and $n$ is the number of nodes); $\sigma_{DC}(\cdot)$ is the activation function of DCNN; $W_{DC}^{t}$ is the weight tensor of DCNN at time step *t*; The dimensions of the hidden state $H_{DC}^t$ at time step *t* and input data $X_{DC}^t$ are the same; and the degree-normalized transition matrix $P_{DC}$ can be obtained by from the adjacency matrix $P_{DC} = D^{-1}A$. Simulation results have shown that DCNN is not only suitable for graph classification, but can also be applied to edges classification tasks, which require augmenting the adjacency matrix and transforming edges to nodes.

Furthermore, a dual-graph convolutional network (DGCN) that consists of two graph convolutional layers in parallel is proposed to jointly account for the global consistency and local consistency in [32]. The first graph convolutional layer is the same as Eq. (8). The second graph convolutional layer substitutes the positive pointwise mutual information (PPMI) matrix for the adjacency matrix:

$$H_{DG}^t = \sigma_{DG}\left(D_P^{-\frac{1}{2}} X_P D_P^{-\frac{1}{2}} H_{DG}^{t-1} W_{DG}^t\right) \quad (12)$$

where $D_P$ is the diagonal degree matrix of PPMI matrix $X_P$; $\sigma_{DG}(\cdot)$ is the activation function of DGCN; $H_{DG}^t$ is the output data of DGNN at time step *t*; and $W_{DG}^t$ is the weight tensor of DGNN at time step *t*.

3) Comparison between spectral and spatial-based GCNs. The main differences between spectral-based GCNs and spatial-based GCNs are as follows:

Firstly, spectral-based GCNs either need to deal with the whole graphs simultaneously or perform eigenvector computation, which leads to more computations induced by the forward and inverse graph Fourier transforms [33]. Relatively, spatial-based GCNs are extensible to large-scale graphs, since they directly define convolutional operations in graph domains. The computation of spatial-based GCNs can be performed in a batch of nodes in place of the whole graph-structured data.

Secondly, spectral-based GCNs which rely on the Fourier transform generalize unfavorably to various graphs. Any perturbations in the graph-structured data will cause the eigenbasis to change, because they assume that the graphs are fixed [34]. In contrast, spatial-based GCNs perform graph convolutional operations locally on each node, and the weights of networks can be easily shared across different locations.

Thirdly, most of spectral-based GCNs are limited to handle undirected graphs [12], while spatial-based GCNs are more flexible to handle multisource graphs, such as directed graphs [31], heterogeneous graphs [35], edge inputs [36], and signed graphs [37], since these graphs can be easily incorporated into the aggregation function.

In general, spectral-based GCNs perform convolutional operations in spectral domains through the complex Fourier transform, while spatial-based GCNs directly define convolutional operations in graph domains. Therefore, spatial-based GCNs show stronger generalization and flexibility compared with most of spectral-based GCNs. In fact, both the spatial-based and spectral-based GCNs are developing unceasingly. For example, Ref. [34] is a kind of spectral-based GCNs that is much efficient already. Therefore, the effectiveness of spectral-based GCNs is also being developed.

### C. Graph Recurrent Neural Networks

The GRNNs are designed for problems defined in graph domains (e.g., classifications in graph-level and node-level) which require outputting sequences. This section will introduce two popular variants of the GRNNs.

To solve long-term dependencies in the graph-structured data and reduce the restrictions in GNNs, there is increasing interest in extending gate mechanisms from RNNs, such as gated recurrent units (GRUs) and long short-term memories (LSTMs)[38], to GRNNs. For example, the graph GRUs are introduced into the propagation step in [39]. Specifically, the gated GNNs expand the RNNs to a fixed number of *T* steps, and calculate the gradients by back-propagating time. The basic mathematical formula of the propagation is

$$\begin{cases} a_v^t = A_v^T \left[h_1^{t-1}, \cdots, h_n^{t-1}\right]^T + b \\ z_v^t = \sigma_Z\left(W^z a_v^t + U^z h_v^{t-1}\right) \\ r_v^t = \sigma_R\left(W^r a_v^t + U^r h_v^{t-1}\right) \\ \hat{h}_v^t = \tanh\left(W_v^t a_v^t + U_v^t\left(r_v^t \odot h_v^{t-1}\right)\right) \\ h_v^t = \left(1 - z_v^t\right) \odot h_v^{t-1} + z_v^t \odot \hat{h}_v^t \end{cases} \quad (13)$$

where $A_v$ is a part of the adjacency matrix, which is used to represent the connection between node *v* and its neighbors; *b* is the offset vector; $W^z$ and $U^z$ are the weights of the update gate; $W^r$ and $U^r$ are the weights of the reset gate; $W_v^t$ and $U_v^t$ are the weights of the gated GNNs at time step *t*; $z_v^t$ is the update gate; $\sigma_Z(\cdot)$ is the activation function of the update gate; $\sigma_R(\cdot)$ is the activation function of the reset gate; $h_v^t$ is the hidden state at time step *t*; and $r_v^t$ is the reset gate. These gates like update functions of GRUs which combine the previous time step and information of other nodes to update the hidden state of each node.

Similar to GRUs, LSTMs are also a popular framework for improving the effectiveness of long-term information propagation. The basic LSTMs architecture called the child-sum tree-LSTM is proposed in [40]. It includes the input gate $i_v$, memory unit $c_v$, output gate $o_v$, and hidden state $h_v$. In addition, it replaces the single forget gate with a forget gate $f_{vk}$ for each child *k*, which results in node *v* aggregate information from its child nodes accordingly. Its mathematical formula is:

$$\begin{cases} \hat{h}_v^{t-1} = \sum_{k \in N(v)} h_k^{t-1} \\ i_v^t = \sigma_I\left(W^i X_v^t + U^i \hat{h}_v^{t-1} + b^i\right) \\ f_{vk}^t = \sigma_F\left(W^f X_v^t + U^f \hat{h}_k^{t-1} + b^f\right) \\ o_{vk}^t = \sigma_O\left(W^o X_v^t + U^o \hat{h}_k^{t-1} + b^o\right) \\ u_v^t = \tanh\left(W^u X_v^t + U^u \hat{h}_k^{t-1} + b^u\right) \\ c_v^t = i_v^t \odot u_v^t + \sum_{k \in N(v)} f_{vk}^t \odot c_k^{t-1} \\ h_v^t = o_v^t \odot \tanh\left(c_c^t\right) \end{cases} \quad (14)$$

where $X_v^t$ is the input data of node $v$ at time step $t$; $\sigma_I(\cdot)$ is the activation function of the input gate; $\sigma_F(\cdot)$ is the activation function of the forget gate; $\sigma_O(\cdot)$ is the activation function of the output gate; $W^i$ and $U^i$ are the weights of the input gate; $W^f$ and $U^f$ are the weights of the forget gate; $W^o$ and $U^o$ are the weights of the output gate; $W^u$ and $U^u$ are the weights of the child-sum tree-LSTM; $b^i$ is the bias vector of the input gate; $b^f$ is the bias vector of the forget gate; $b^o$ is the bias vector of the output gate; $b^u$ is the bias vector of the child-sum tree-LSTM; and $h_v^t$ is the hidden state at time step $t$.

### D. Graph Attention Networks

In the above-mentioned GCNs, the neighborhood of nodes is aggregated with equal or predefined weights. Nevertheless, the impacts of neighbors may vary greatly [41]. Therefore, they should be learned in the process of training, not predetermined. Activated by attention mechanisms, GATs introduce the attention mechanism into graph domains by revising the graph convolutional operation [42]:

$$H_{\text{GATs},i}^{t+1} = \sigma_{\text{GATs}}\left(\sum_{j \in N(i)} \alpha_{ij}^t W_{\text{GATs}}^t H_{\text{GATs},j}^t\right) \quad (15)$$

where $H_{\text{GAT},i}^{t+1}$ is the hidden state of node $i$ of GATs at time step $t+1$; $N(i)$ is the neighborhood of node $i$; $W_{\text{GATs}}^t$ is the weights of GATs at time step $t$; $\sigma_{\text{GATs}}(\cdot)$ is the activation function of GATs; and $\alpha_{ij}^t$ is the attention coefficient of node $j$ to node $i$ at time step $t$. Its mathematical formula is:

$$\alpha_{ij}^t = \sigma_{\text{SM}}\left(e_{ij}^t\right) = \frac{\exp\left(e_{ij}^t\right)}{\sum_{k \in N(i)} \exp\left(e_{ik}^t\right)} \quad (16)$$

$$e_{ij}^t = \sigma_{\text{LR}}\left(F\left[W_{\text{GATs}}^t H_{\text{GATs},i}^t \| W_{\text{GATs}}^t H_{\text{GATs},j}^t\right]\right) \quad (17)$$

where $\|$ is the concatenation operation; $\sigma_{\text{LR}}$ is the leaky rectified linear unit (LeakyReLU) function; $\sigma_{\text{SM}}$ is the Softmax function; and $F$ is a function (e.g., multi-layer perceptron) to be learned.

Furthermore, the multi-head attention mechanism is utilized to stabilize the learning process. The $K$ independent attention mechanisms are used to calculate hidden states and then concatenate these features in graph attention network (GAT) [43], which lead to two different output representations:

$$H_{\text{GAT1},i}^{t+1} = \|_{k=1}^{K} \sigma_{\text{GAT1}}\left(\sum_{j \in N(i)} \alpha_{ij}^k W_{\text{GAT},k}^t H_{\text{GAT},j}^t\right) \quad (18)$$

$$H_{\text{GAT2},i}^{t+1} = \sigma_{\text{GAT2}}\left(\frac{1}{K}\sum_{k=1}^{K}\sum_{j \in N(i)} \alpha_{ij}^k W_{\text{GAT},k}^t H_{\text{GAT},j}^t\right) \quad (19)$$

where $\|$ is the concatenation operation; $\sigma_{\text{GAT1}}$ is the activation function of the first GAT; $\sigma_{\text{GAT2}}$ is the activation function of the second GAT; $H_{\text{GAT1},i}^{t+1}$ is the hidden state of node $i$ of the first GAT at time step $t+1$; $H_{\text{GAT2},i}^{t+1}$ is the hidden state of node $i$ of the second GAT at time step $t+1$; and $\alpha_{ij}^k$ is normalized attention coefficient calculated by the $k^{\text{th}}$ attention mechanism.

Another popular variant is the gated attention network (GaAN) which employs the dot product attention and key-value attention mechanism [44]. To replace the average operation, the self-attention mechanism is employed to gather information from different heads.

### E. Graph Generative Networks

The purpose of GGNs is to generate some new graph-structured data by learning a series of given historical samples. Similar to generative networks in Euclidean domains, the existing GGNs mainly include the graph automatic encoders (GAEs), variational graph auto-encoders (VGAEs), and graph generative adversarial networks (GGANs) [45].

The GAEs consist of an encoder and a decoder [46]. Firstly, the features $X$ and the adjacency matrix $A$ of the nodes are fed to the encoder to obtain the embedding matrix $Z_{\text{GAE}}$ of the graph-structured data:

$$Z_{\text{GAE}} = \text{GCN}(X, A) \quad (20)$$

Then, the decoder of GAEs aims to reconstruct the graph adjacency matrix by feeding the embedding matrix $Z_{\text{GAE}}$ from the encoder:

$$\tilde{A} = \sigma_{\text{GAE}}\left(Z_{\text{GAE}} Z_{\text{GAE}}^T\right) \quad (21)$$

where $\tilde{A}$ is the reconstructed adjacency matrix; and $\sigma_{\text{GAE}}(\cdot)$ is the activation function of GAEs.

The new graph-structured data generated by GAEs lacks diversity and the number is limited. To overcome these shortcomings, VGAEs introduce probability to GAEs [47]. The loss function of VGAEs is the variational lower bound:

$$\mathcal{L} = \mathbb{E}_{q(Z_{\text{VG}}|X,A)}\left[\log p(A | Z_{\text{VG}})\right] - \text{KL}\left[q(Z_{\text{VG}} | X, A) \| p(Z_{\text{VG}})\right] \quad (22)$$

where $\text{KL}[\cdot]$ denotes the Kullback-Leibler divergence; $p(Z_{\text{VG}})$ is the Gaussian distribution; $Z_{\text{VG}}$ is the Gaussian noise; $p(A | Z_{\text{VG}})$ is the inner product between latent variables; $q(Z_{\text{VG}} | X, A)$ is empirical distribution of nodes, which is used to approximates the prior distribution. VGAE can only approximate the lower bound of logarithm likelihood of the nodes, which results in the limited quality of the new graph-structured data.

In order to improve the quality of the generated graph-structured data, the adversarial loss function of GGANs is introduced into the training process [48]. Given the graph $G$, GGANs aim to train the generator and discriminator by historical samples. Specifically, the generator $G(v | v_i; \theta_G)$ attempts to fit the real connected distribution $P_{\text{real}}(v | v_i)$ of the nodes as much as possible and generates the most likely nodes connected with node $v_i$ from the nodes set to deceive the discriminator. On the contrary, the discriminator $D(v, v_i; \theta_D)$ tries to identify the connectivity for the nodes pair $(v, v_i)$ and outputs a value that represents the probability that whether the node is ground-truth neighbors of $v_i$ or the one generated by the generator. Formally, the generator and discriminator are playing the two-player min-max game, and the loss function of GGANs is:

$$\min_{\theta_G} \max_{\theta_D} L = \sum_{i=1}^{n} \left( E_{v \sim P_{\text{real}}(\cdot|v_i)} \left[ \log D(v, v_i; \theta_D) \right] + E_{v \sim G(\cdot|v_i; \theta_G)} \left[ \log(1 - D(v, v_i; \theta_D)) \right] \right) \quad (23)$$

where $n$ is the total number of nodes; $\theta_D$ is the parameters of the discriminator to be learned; and $\theta_G$ is the parameters of the generator to be learned. The parameters of the generator and discriminator are updated during the training process by alternately maximizing and minimizing the loss function.

*F. Spatial-Temporal Graph Neural Networks*

Normally, the structure and feature information of many graph-structured data may change with time. For example, the power curves of adjacent wind farms have temporal-spatial correlation, which will change with times and environmental factors, such as wind speed and wind direction. There is a need to consider spatial dependence when forecasting the output powers of wind farms [49]. To predict values or graph labels of nodes, the STGNNs are designed to capture temporal-spatial dependencies of graphs simultaneously. Existing STGNNs mainly include two categories: RNNs-based models and CNNs-based models.

For the RNNs-based models, they try to capture temporal-spatial dependencies by filtering hidden states and input data passed to recurrent units using graph convolutional operations [44]. Normally, the mathematical formula of RNNs is shown as follows:

$$H_{\text{RNN}}^t = \sigma_{\text{RNN}} \left( W_{\text{RNN}}^t X_{\text{RNN}}^t + U_{\text{RNN}}^t H_{\text{RNN}}^{t-1} + b_{\text{RNN}}^t \right) \quad (24)$$

where $W_{\text{RNN}}^t$ and $U_{\text{RNN}}^t$ are the weights of RNNs; $b_{\text{RNN}}^t$ is the offset vector; $X_{\text{RNN}}^t$ is the feature matrix of nodes at time step $t$; $\sigma_{\text{RNN}}(\cdot)$ is the activation function of RNNs; and $H_{\text{RNN}}^t$ is the hidden states of RNNs at time step $t$. After inserting graph convolutional operations, Eq. (24) becomes:

$$H_{\text{RG}}^t = \sigma_{\text{RNN}} \left( \text{GCN}(X_{\text{RNN}}^t, A; W_{\text{RNN}}^t) + \text{GCN}(H^{t-1}, A; U_{\text{RNN}}^t) + b_{\text{RNN}}^t \right) \quad (25)$$

where $H_{\text{RG}}^t$ is the hidden states of RNNs-based models at time step $t$; and $A$ is the adjacency matrix.

On this basis, the diffusion convolutional RNNs combine a GRUs with diffusion graph convolutional layers in [50], while the graph convolutional recurrent networks (GCRNs) incorporates the LSTMs into ChebNet in [51].

In addition, some works utilize edge-level RNNs and node-level RNNs to deal with different aspects of temporal features [52]. For example, a recurrent structure with the edge-level RNNs and the node-level RNNs is proposed to forecast labels of nodes at each time step in [53]. The temporal information of nodes and edges passes through the edge-level RNNs and node-level RNNs respectively. The output of the edge-level RNNs is used as the input data of the node-level to merge spatial information.

The RNNs-based models have gradient vanishing problems and time-consuming iterative propagation. Relatively, the CNNs-based models handle the temporal-spatial graphs in a non-recursive manner, which has the advantages of the stable gradient, parallel computing, and low memory requirement [54]. For example, A PGC layer and a 1-D convolutional layer are used to build temporal-spatial block in [55]. The CGCNs [56] combine one-dimensional convolutional layers with ChebNet, which builds a temporal-spatial block by sequentially integrating graph convolutional layers and gated one-dimensional convolutional layers.

*G. Hybrid forms of GNNs*

In addition to the GNNs mentioned above, there are also some extended models, such as GRL [57] and GTL [58].

The combination of RL and GNNs has led to a new research field named GRL, which integrates the decision-making of RL and perception of GNNs. Therefore, GRL can be applied to a variety of tasks requiring both the precise control and rich perception of graph-structured data. Many researches on GRL are currently being conducted. For example, to predict chemical reaction products, a graph transformation policy network is proposed in [59], which combines the strengths of RNNs and RL to learn the chemical reactions directly from raw data with minimal prior knowledge. Specifically, it uses the RNNs to memorize the forecasting sequences and the GCNs to learn the representations of nodes. In [60], a graph convolutional policy network that consists of GCNs and the RL is proposed to discover novel molecules with specific properties. The generative network is considered as an RL agent performing in the graph generative environment. In addition, the graph generation is regarded as a Markov decision process, where edges and nodes are added. Similarly, the GGANs are proposed to directly operate on small molecular graphs in [61]. It combines a reinforcement learning objective with GNNs to encourage the generation of small molecular graphs with specific desired chemical properties.

GTL is a research topic in machine learning, which aims to store prior knowledge when solving old problems and applying it to new but related problems [62]. Although GNNs have shown superior performance in various fields, training dedicated GNNs will be costly for large-scale graphs. In this case, a practically useful and theoretically grounded framework is proposed for the transfer learning of GNNs in [63]. The proposed novel views towards the important graph information and activate the capturing of it as the goals of transferable GNNs training, which motivates the design of GNNs frameworks. In [64], a model containing transfer learning and GNNs is proposed to solve related tasks in the target domain without training a new model from scratch by transferring the natural geometric information learned in the source domain.

III. APPLICATIONS IN POWER SYSTEMS

After years of development, some papers about the application of GNNs have been published, and most of them have been published since 2018. However, most of these applications are focused on computer science and biology, such as social networks, link prediction, protein structure generation, and natural language processing. The applications of GNNs in power systems are relatively limited. Table II lists some existing publications of GNNs in power systems, covering fault scenario application, time series prediction of RES and loads, power flow calculation, data generation, etc. In the future, it may be divided into more categories as the number of publications increases. To serve as a catalyst for further study of applications, this section reviews these available literatures and discusses oriented researches.

TABLE II
EXISTING APPLICATIONS OF GRAPH NEURAL NETWORKS

| Area | Category | Involved GNNs | Ref |
|---|---|---|---|
| Fault scenario application | Transformer fault diagnosis | Spectral-based GCNs | [67] |
| | Fault location | Spectral-based GCNs | [69],[70] |
| | Fault detection and isolation | Spectral-based GCNs | [72],[76] |
| | Power outages prediction | Spectral-based GCNs | [74] |
| Time series prediction | Solar power prediction | Spectral-based GCNs | [78]-[81] |
| | Wind power or wind speed prediction | Graph LSTMs and Spectral-based GCNs | [82]-[84] |
| | Residential load prediction | Spectral-based GCNs | [85] |
| Power flow calculation | Power flow approximation | Spatial-based GCNs, GAEs | [86]-[88][90] |
| | Optimal power flow | Spectral-based GCNs | [6],[91] |
| | Optimal load shedding | Spectral-based GCNs | [92] |
| Data generation | Scenario generation | Spectral-based GCNs and GAEs | [93],[94] |
| | Synthetic feeder generation | Spectral-based GCNs and GGANs | [95] |
| Others | Coupled power and transportation networks | Spectral-based GCNs and GRL | [96] |
| | Line flow control | Spectral-based GCNs and GRL | [97] |
| | Maintenance tasks | Hybrid forms of GNNs | [98] |
| | Operation of distributed energy resources | Spectral-based GCNs | [99], [100] |
| | Safe operation of power grids | Spectral and spatial-based GCNs | [101]-[103] |
| | Synchrophasors recovery | Graph LSTMs, Spectral-based GCNs and GGANs | [104] |
| | Transient stability assessment | Graph LSTMs, Spectral-based GCNs | [105] |
| | Wind power estimation | Physics-induced GNNs | [106] |

*A. Fault scenario application*

In this section, the four categories applications given under fault scenario and some potential research direction are discussed. The main aspects in this relation are to detect failures in order to avoid power outages and ensure the safe operation of the power system.

*A1. Transformer fault diagnosis*

One often used diagnostic method within transformer fault diagnosis is by applying dissolved gas analysis. the fault diagnosis of the dissolved gas analysis data is crucial to diagnose the incipient faults of power transformers as early as possible. Most of the existing methods for transformer fault diagnosis can be classified into the following three categories: model-based methods (e.g., CNNs), distance-based methods (e.g., k-nearest neighbors), and their hybrid forms (e.g., ensemble models). Specifically, distance-based methods attempt to calculate the similarity metrics between and the historical samples and the samples with unknown labels [65], but they are hard to mine the complex non-linear relationship between dissolved gas data and labels. In contrast, model-based methods predict the type of fault through a classifier trained with historical samples [66], while they ignore the similarity metrics or do not use them directly. Furthermore, the hybrid model can not only account for similarity metrics, but also train a classifier for fault diagnosis. GCN can be considered as a kind of hybrid form. On one hand, similarity metrics between labeled samples and unknown samples can be represented by an adjacency matrix. On another hand, GCNs can accurately explore the complex relationship between dissolved gas data and fault types through graph convolutional layers. As shown in Fig. 3, the spectral-based GCNs are proposed to improve the accuracy of transformer fault diagnosis in [67]. The simulation results show that the performance of the GCNs is better than those of the traditional methods, such as multi-layer perceptions (MLPs), support vector machines (SVMs), CNNs, and k-nearest neighbors (KNNs) in different data volumes and input features. Although GCNs show strong performance in transformer fault diagnosis, there are still some potential directions for further research: 1) Since the size of the adjacency matrix depends on the number of samples, GCNs need to be retrained when the number of samples changes, which lead to that the existing GCNs methods are difficult to use for on-line fault diagnosis. It needs further study on how to avoid retraining GCNs for transformer fault diagnosis. For example, it may be considered that new samples and their most similar samples have the same connection relationship with others. In this case, GCNs can identify new samples without repeated training. 2) The existing paper only analyzes the performance of spectral-based GCNs for transformer fault diagnosis. This work may be extended to spatial-based GCNs and then explore the performances of different GCNs for transformer fault diagnosis.

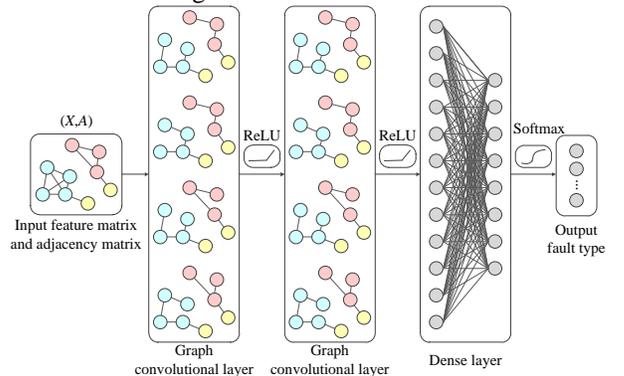
Fig. 3. Transformer fault diagnosis via the spectral-based GCNs [67].

*A2. Fault location*

Traditional methods of fault location for distribution networks mainly include voltage sag-based methods, impedance-based methods, traveling wave-based methods, and auto-mated outage mapping methods. Although they have their own advantages, there are two main challenges [68]: Firstly, they cannot flexibly combine measurement data from different buses, especially in the case of data loss. Secondly, most of traditional methods have difficulties in modeling the topology of the distribution network. In order to solve these problems, Ref. [69] employs the graph GRUs to automatically localize the faults of distribution networks. The feeder topology is represented by graph edges and problem data (e.g., measurements and electrical characteristics) is regarded as graph nodes. Similarly, the spectral-based GCNs are proposed to explore comprehensive information from multiple measurement units and capture the spatial correlations among buses in [70]. As shown in Table III, simulation results show that classification accuracy and one-hop accuracy of GCNs are higher than those of some machine learning methods, such as the hybrid model of the principal component analyses (PCAs) and SVMs, hybrid model of PCAs and random forests (RFs), and MLPs. Furthermore, there are still some potential directions for further research: 1) The effectiveness of GCNs in more realistic settings needs to be further studied. For example, field data can be utilized to fine-tune the pre-trained model through graph transfer learning. Especially trained GCNs may be transferred to other distribution networks with different topologies. 2) The difference between spectral-based GCNs and spatial-based GCNs on the performance of large-scale distribution networks can be analyzed, and the impact of different RES on GCNs can be discussed.

TABLE III
FAULT LOCATION ACCURACIES OF DIFFERENT METHODS IN THE IEEE 123 BUS SYSTEM

| Model | Accuracy | One-hop accuracy |
|---|---|---|
| PCAs+SVMs | 94.60% | 98.31% |
| PCAs+RFs | 94.96% | 99.28% |
| MLPs | 84.64% | 96.38% |
| GCNs | 99.26% | 99.93% |

*A3. Fault detection and isolation*

Traditional methods for fault detection and isolation are used to identify and isolate faults on the level of a single component by accounting for features from this component and corresponding components [71]. These methods are not good enough, since they are independently applied to a single component without explicitly considering the dependencies among multiple components coexisting in power systems. The interaction between components brings challenges to fault isolation. In addition, the traditional methods do not consider the network structure when designing the fault diagnosis, which causes over-fitting problems. To solve these problems, the connected components in power systems can be represented as a weighted undirected graph structure [72]. Then, local relationships between power variables in different components of the distribution network are explored by GCNs to improve fault detection and isolation. Simulations show that GCNs are significantly better than several baselines. Moreover, the community-varying GCNs can be used to explore highly correlated components in distribution networks, so as to improve performance for fault detection and isolation.

*A4. Power outages prediction*

Power outages have an important impact on economic development because of the strong correlation between power energy and productive sectors. Traditional methods ignore the connection relationship of measurement data, resulting in their limited accuracy [73]. In order to improve the accuracy of predicting power outages, a new method based on GNNs is proposed to process weather measurements [74]. Specifically, the structure of weather stations is regarded as a graph where the edges denote the distance between these stations, and each node represents a station. Then, weather measurements at stations are modeled as features of each node, and the corresponding topology is utilized to process these features. The simulation results show that GNNs significantly improve the accuracy of power outage prediction. Moreover, this framework may be extended to communities. Each community is modeled as a node, and the connection relationship between communities is modeled as branches. Then, the correlation of power between communities can be further analyzed.

As a further example of power outages, a case with a PV plant is considered. Generally, PV plants are installed in remote places where the weather may be very bad, which leads to PV plants failure is difficult to predict. Traditional methods either have low accuracy or require a large number of historical samples to train the classifier, which is not suitable for photovoltaic fault classification [75]. Therefore, a graph signal processing technique is proposed to detect PV faults with a limited amount of labeled samples [76]. The simulation results show that the graph-based classifier has higher accuracy and lower computational cost than traditional methods, such as KNNs, SVMs, and random forests. In Euclidean domains, Siamese networks and matching networks show good performance for few-shot learning. Similarly, it can also employ GCNs or GRNNs to construct graph Siamese networks, which may be suitable for fault diagnosis of the graph-structured dataset with small samples.

*A5. Further ideas for fault scenario application*

In addition to the above applications of fault scenarios, there are some potential directions worthy of further study: 1) Existing works show that ensemble learning can significantly improve predictive performance by using multiple learning algorithms. Therefore, it may ensemble multiple GNNs or combine GNNs with traditional DNNs in the future. 2) In Euclidean domains, the pooling operation of CNNs loses a lot of feature information of input data, which limits the accuracy of fault diagnosis. In order to solve this problem, the capsule network with primary capsule layers and digital capsule layers is proposed to significantly improve the accuracy of fault diagnosis. Similarly, the capsule network may also be extended to graph domains. 3) For GCNs in [67], their adjacency matrix represents the similarity metrics between samples, and the graph convolutional layer captures the complex nonlinear relationship between features of samples and labels. This framework may be suitable for transformer fault diagnosis and other classification tasks in the power system, such as power quality disturbance classification and transient stability assessment of power systems.

*B. Time series prediction*

In this section, the three categories applications given under time series prediction and some potential research direction are discussed.

*B1. Solar power prediction*

Accurately predicting short-term powers is of great significance for power systems with high penetration of RES, because PV plant and wind turbine have a great impact on the economic and stable operation of power systems. However, traditional methods cannot accurately capture the temporal and spatial correlations of photovoltaic stations [77], because high-dimensional input vectors require a large number of free parameters in traditional machine learning models, while the gradient descent method based on the loss function cannot be effective to adjust a large number of free parameters. In [78], the CNNs-based STGNNs are proposed to leverage spatial-temporal coherence among PV systems. While in [79], the GAEs are employed to capture the spatial-temporal manifold of power loads and PV power, which are considered as spatial-temporal graphs representing the measurements of units via nodes and reflecting the mutual correlation between the units via edges. Similarly, the convolutional GAEs are devised to predict probabilistic solar irradiance in different multiple measurement sites, which are modeled as an undirected graph [80]. In order to improve the accuracy of photovoltaic prediction, a hybrid algorithm with the LSTMs and GCNs is proposed in [81]. Specifically, the LSTMs are employed to extract the temporal features of the photovoltaic power curves, and GCNs are used to capture the spatial correlation between multiple adjacent PV plants. Each PV plant is regarded as a node, and the features of each node include historical power and weather data. If the correlation coefficient between the two PV plants is greater than the threshold, they are considered to be connected. As shown in Table IV, the mean absolute error (MAE), mean absolute percentage error (MAPE), root mean squared error (RMSE) of the proposed hybrid model are smaller than those of traditional methods, such as the LSTMs and MLPs. Although this model shows strong performance in short-term prediction of PV power, there are still some potential directions for further research: 1) More combinations of algorithms can be tried. For example, if the performance of the GRUs to capture temporal correlation is similar to that of LSTMs, and the computing time is less than that of LSTMs. The performance of the combination of the GRUs and GCNs for short-term photovoltaic power can be explored in the future. 2) The RNNs in Euclidean domains have been extended to the GRNNs in graph domains, and have shown outstanding performance in natural language processing. Perhaps the temporal and spatial correlation of the PV plants can be captured by GRNNs.

TABLE IV
DAY-AHEAD FORECASTING RESULTS OF DIFFERENT METHODS

| Models | MAPE | RMSE | MAE |
|---|---|---|---|
| LSTMs+GCNs | 28.30% | 1.10 | 0.79 |
| LSTMs | 54.17% | 2.19 | 1.58 |
| MLPs | 43.35% | 1.78 | 1.31 |

*B2. Wind power and wind speed prediction*

Compared with onshore wind powers, the fluctuations and intermittence of offshore wind powers are stronger, which poses great challenges to the operation and planning of power systems. Traditional methods are difficult to fully exploit the spatial property of offshore wind powers, resulting in low forecasting accuracy. To simultaneously represent the temporal and spatial information of wind farms, a superposition graph neural network (SGNN) is proposed in [82], which refers to the superposition structure of CNNs on feature channel and the feature transfer method of GNNs. Specifically, many nearby wind turbines are form of the wind farms. They are considered as the graph-structured data that represent the direct spatial property among wind turbines. To the temporal-spatial property of wind farms, the encapsulated data structure is designed by overlaying spatial maps at different nodes and time horizons. Then, a SGNN is employed to extract features, so as to maximize the utilization of temporal-spatial property. The simulation results show that SGNN can accurately capture the temporal and spatial features of wind farms and achieve higher accuracy than traditional methods. In [83], the spectral-based GCNs are presented to learn the interconnection among multiple wind farms for wind speed prediction. Similarly, a new temporal and spatial wind speed feature learning framework is proposed to combine graph deep learning and rough set theory in [84]. As shown in Fig. 4, the wind farms are modeled as the graph-structured data where nodes with high wind directions and speed correlations are connected by edges. Then, the recurrent LSTMs are used to extract temporal features of each wind site, which are fed to the spectral-based GCNs. To learn robust features, rough set theory is embedded in the GCNs by extracting interval lower-bound and upper-bound filtering parameters. Simulation results show that the model has better performance than popular deep learning architectures, such as deep belief networks (DBNs). Furthermore, there are still some potential directions for further research: 1) The SGNNs may be extended to other prediction tasks, such as irregular distribution point clouds of PV systems. 2) An effective graph construction method is able to accelerate the efficiency of preprocessing in extracting spatial characteristics. Nevertheless, The SGNNs only use a feasible construction method without involving more theories about computer graphics, which can be further studied.

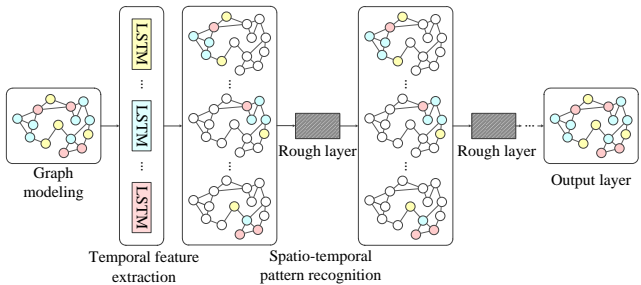

Fig. 4. Structure of spatio-temporal graph deep neural network for short-term wind speed prediction [84].

*B3. Residential load prediction*

Short-term load prediction is an important part of providing a stable power supply to all electricity consumers on power systems. In order to accurately predict household loads, a novel method that consists of graph spectral clustering and non-intrusive load monitoring is proposed in [85]. Specifically, the aggregated power curves are decomposed into the power curve of the individual appliance which is forecasted separately. Then, the total power curve is obtained by aggregating the forecasted power curve of individual appliances. Every electrical appliance is regarded as a node, and branches are constructed by functions of state duration probabilities of appliances. Simulation results show that this model is more accurate compared to existing approaches, such as similar profile load forecast and autoregressive integrated moving average.

*B4. Further ideas for time series prediction*

In addition to the above applications of prediction for RES and loads, there are some potential directions worthy of further study: 1) In Euclidean domains, CNNs and LSTMs are often combined to predict powers of RES or power loads, since CNNs are better at automatically extracting the features of input data and representing the complex nonlinear relationship between the features and the real powers, while LSTMs are better at capturing the temporal correlation of time series. Therefore, they are often used to construct hybrid models. In the same way, it may also combine GCNs and GRNNs to construct a hybrid model with their advantages in graph domains. 2) Temporal convolutional network (TCN) is a novel neural network originated from the 1-dimensional CNNs. It keeps the powerful ability of feature extraction of CNNs, and is very suitable for forecasting time series. TCN may also be extended to graph domains, and test its performance in load and renewable energy power prediction. 3) Like fault diagnosis, ensemble learning can also be used to improve the forecasting performance of GNNs. It may even ensemble the traditional RNNs and GRNNs, as well as the impact of different ensemble frameworks on the accuracy.

*C. Power flow calculation*

The power flow calculation of the power system is the basis for operation and planning. In this section, the three categories applications given under power flow calculation and some potential research direction are discussed.

*C1. Power flow approximation*

Due to the growing integration of RES and an increasing amount of power equipment, the physical models of power systems are becoming more complex, which leads to the longer computing time of power flow calculation. This requires further development of fast power flow approximation methods. Traditional methods (e.g., MLPs and CNNs) do not exploit the intrinsic network topology of power systems, resulting in low accuracy. Therefore, the spatial-based GCNs are employed to approximate the power flow in [86]. Specifically, the active power and reactive power of each node are regarded as features, and the power flow of each branch and the voltage of each node are defined as corresponding tags to be predicted. Simulation results show that the spatial-based GCNs provide a highly accurate power flow approximation and outperform classical methods on large-scale power systems. In [87], the voltage magnitude and voltage angle are estimated by the spectral-based GCNs. In [88], the ChebNet is employed to calculate distribution characteristics of power flow without the prior knowledge. The GCNs show shorter computation time and higher accuracy than the conventional Monte-Carlo method. To decompose and solve the power flow equations of transmission networks, the spectral-based GCNs are presented to provide a geometric picture of the electrical variables in [89]. Similarly, to calculate the power flow quickly and in parallel, the GNNs are proposed to minimize the violation of Kirchhoff's law at each node during training in [90]. Unlike traditional methods, this graph neural solver learns by itself and does not imitate the output data of the Newton-Raphson solver. Simulation results show that the GNNs can perform predictions faster than traditional methods, such as the Newton-Raphson solver. Although this model shows strong performance in power flow calculation of power systems, there are still some potential directions for further research: 1) The distribution networks are constantly changing. When the nodes and branches of the distribution networks increase, the trained models cannot be used directly. In the next step, it can study how to fine-tune the trained model through transfer learning so that the model can be applied to the expanded distribution networks. 2) Alternative GCNs model and further investigation about architecture improvements may be considered in the future. 3) In Euclidean domains, adversarial training has been successfully applied to regression tasks. The adversarial training, such as GGANs may also be introduced into the power flow calculation.

*C2. Optimal power flow*

In addition to being used for power flow calculation, GNNs can also be further applied to the optimal power flow of distribution networks. For example, the spectrum-based GCNs are designed to optimize the reactive power of distribution networks in [6]. Specifically, the adjacency matrix is used to represent the topology information between the nodes in distribution networks, so as to mine the correlation of nodes. Then, the deep graph convolutional layer is used to capture the complex nonlinear relationship between the state of the power equipment and the power loads. Simulation results show that the performance of this model is better than those of traditional data-driven methods, such as CNNs, MLPs, and case-based reasoning. Similarly, the GNNs are designed to approximate the optimal power flow solution in [91]. GNNs are local information and scalable processing architectures that mine the network structure of the input data. It is trained by taking a given network state as input and using the output results to approximate the optimal solution of interior-point optimizer. Simulation results show that local solutions adequately exploit the latent grid structure and outperform other comparable methods. Furthermore, there are still some potential directions for further research: 1) The existing GNNs are difficult to account for the constraints of optimal power flow, such as voltage and current constraints, which may cause power grids to operate in an unsafe state. How to consider the constraints in GNNs can be further studied. 2) The existing methods assume that the topology of power systems is invariant, i.e. the adjacency matrix is fixed. However, the reconfiguration of the distribution network is also a means of regulating power flow. The distribution network can be regarded as a kind of spatial-temporal graph, the adjacency matrix and features of nodes change with times. In this case, none of the existing GCNs can be directly applied to optimal power flow calculations. Furthermore, spatial-temporal graph convolutional networks have shown outstanding performance in the field of computer vision for spatial-temporal graphs. Therefore, they may be extended to the optimal power flow of distribution networks. 3) The existing methods are only suitable for solve static optimal power flow, which is difficult to be directly used in actual engineering due to the fluctuation of power loads. GCNs may be extended to the dynamic optimal power flow.

*C3. Optimal load shedding*

Load shedding is very important for operations of distribution networks particularly under contingency events, such as line failure. Since traditional methods need to solve complex optimization models, the computing time is very long, which cannot meet the real-time requirements of distribution networks. To solve these problems, the GCNs are used for load-shedding operations in [92]. Specifically, the features of nodes include voltage amplitude, voltage angle, active power, and reactive power. The adjacency matrix represents the topology information of distribution networks. As shown in Table V, this model significantly outperforms the MLPs and linear regressions (LRs) in different testing systems. Furthermore, the existing methods do not account for the impact of graph pooling layers and dropout layers on accuracy and computing time, which can be further discussed.

*C4. Further ideas for power flow calculation*

In addition to the above applications of power flow calculation, there are some potential directions worthy of further study: 1) In Euclidean domains, DBNs also show good performance in the optimal power flow of distribution networks. DBNs may be generalized to graph domains for the optimal power flow. 2) The existing methods need massive data to train GNNs. The next step is to design a structure suitable of GNNs for the dataset with small samples. Besides, GGANs may also be used to expand samples to train these GNNs.

TABLE V
PREDICTION RESULTS OF DIFFERENT TESTING SYSTEMS

| System | Model | RMSE(Training) | RMSE(Testing) |
|---|---|---|---|
| 9-bus | GCNs | 0.0282 | 0.0685 |
| | MLPs | 0.0940 | 0.1209 |
| | LRs | 0.0929 | 0.0980 |
| 30-bus | GCNs | 0.1330 | 0.1457 |
| | MLPs | 0.1375 | 0.1582 |
| | LRs | 0.1446 | 0.1734 |
| 57-bus | GCNs | 0.0744 | 0.1231 |
| | MLPs | 0.1158 | 0.6394 |
| | LRs | 0.1507 | 0.5251 |
| 118-bus | GCNs | 0.0086 | 0.0291 |
| | MLPs | 0.1034 | 6.5755 |
| | LRs | 0.2422 | 3.1741 |

*D. Data generation*

In this section, the two categories applications given under data generation and some potential research direction are discussed.

*D1. Scenario generation*

Stochastic scenario generation is an important means to capture the uncertainties of solar irradiance by generating a set of possible time series. Traditional methods need to artificially assume the probability distributions of solar curves, and use historical samples to fit the key parameters in the probability distribution, which leads to poor quality and generalization ability. To solve these problems, the convolutional GAEs are introduced to capture generate samples from the probability densities learned at each node in [93]. The simulation results show that the generated samples are very similar to the observed historical samples, which is helpful for photovoltaic probability prediction. In [94], the spatial-temporal graphs are used to represent both the temporal correlation of the bus states and the spatial correlation among the buses in distribution networks. Then, a regression neural network is trained to generate missing measurements. Furthermore, there are still some potential directions for further research: 1) The new graph-structured data generated by GAEs lacks diversity and the number of samples is limited. Therefore, GAEs may be replaced with VGAEs which generate any solar irradiance by feeding Gaussian noises. 2) GAEs may be extended to conditional VGAEs where Gaussian noise and labels are fed into the generator to obtain the solar irradiance with specified properties, such as fluctuating patterns.

*D2. Synthetic feeder generation*

To generate new feeders of distribution networks similar to real samples in both topology and attributes, the GGANs with a discriminator and a generator are proposed in [95]. A typical framework of GGANs is shown in Fig. 5. Specifically, the device connectivity and device properties are represented with the adjacent matrix and feature matrix to allow GGANs to learn topology and features from real distribution network feeder model input files. Wasserstein distance is employed to optimize the GGANs for distinguishing the generated graphs from the real samples. A baseline comparison of feeder generation is performed on a dataset, as shown in Table VI [95]. Simulation results show that GGANs improve the connected rate by 18.9%, the perfect rate and success rate by 27.2%. Furthermore, GGANs may be combined with VGAEs to improve the quality of generated samples and avoid vanishing gradients and exploding gradients problems in the training process.

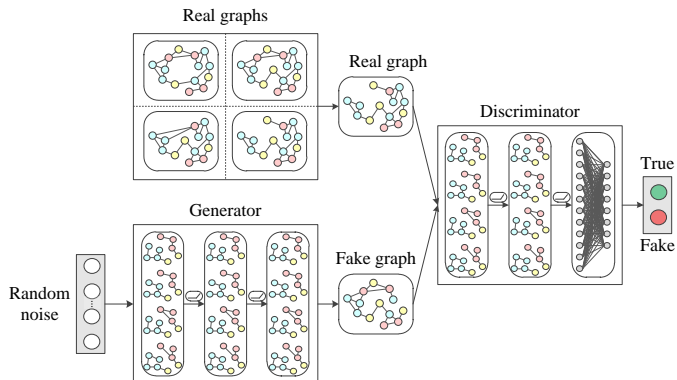

Fig. 5. Framework of the typical GGANs [95].

TABLE VI
PERFORMANCE COMPARISON OF DIFFERENT METHODS

| Indicator | Random matrix-based method | GGANs |
|---|---|---|
| Connected rate | 8.3% | 27.2% |
| Success rate | 0 | 27.2% |
| Perfect rate | 0 | 27.2% |

*D3. Further ideas for data generation*

In addition to the above applications of data generation, there are some potential directions worthy of further study: 1) In Euclidean domains, there are some other deep generative networks with strong performance, such as flow-based generative networks, implicit maximum likelihood estimation models, and generative moment matching networks. They may be extended to graph domains for data generation of power systems. 2) Besides feeder generation, GGNs may also be applied to other fields of power systems, such as modeling power curves for loads and RES.

*E. Others*

There are some other papers that apply GNNs to other technical branches of power systems.

To deal with the huge challenge of large-scale disordered and fast-charging electric vehicles to coupled power transportation networks, a multi-objective optimization model including electric vehicle-charging station, transportation network, and power systems is proposed in [96]. Then, fast charging guidance strategy and regular environment information extraction are realized by using GRL. In addition, GRL can also be used for other control problems in power systems. For example, Ref. [97] proposes a simulation-constraint GRL to provide solutions for line flow control of power systems. The spectral-based GCNs are employed to extract features from network topology and real measurements.

Normally, power communication networks have difficulty in quickly obtaining knowledge and completing the maintenance tasks. To solve these problems, a Relation-Tuple-Entity Heterogeneous GNNs are proposed to capture semantic information in different granularities for knowledge reasoning in [98].

In order to coordinately control distributed energy resources, such as RES and electric vehicles, Ref. [99] proposes a model-free probabilistic graphical framework to predict the grid congestion of power systems using GNNs. In [100], a ChebNet is presented to solve economic dispatch and unit commitment in a day ahead trading market of power systems.

The automatic topology identification of distribution networks is crucial for the data-driven safe operation of power grids. In [101], the input data is represented as the graph-structured data, and then the spectral-based GCNs are employed to identify power network topology. In [102], the novel hybrid forms of GNNs are designed to test if medium voltage distribution networks satisfy the safe property of the topology or not. To improve the state awareness of distribution networks, Ref. [103] uses the spatial-based GCNs to realize the super-resolution of measurements, such as topology graphs.

Data integrity of power systems plays a crucial role in the operation and control of smart grids, since state measurements and communication latency are not immediately available at the control center, resulting in slow responses of time-sensitive applications. In [104], a graph convolutional recurrent adversarial network is designed to extract graphical information and temporal correlations of data.

Reliable online transient stability assessment is critical to the safe operation of power systems. To improve the accuracy of transient stability assessment, Ref. [105] aggregates the LSTMs and GCNs to obtain a recurrent graph convolutional network where the LSTMs subsequently capture the temporal features and GCNs integrate the node states with the topological information.

To accurately estimate power outputs of wind turbines in different conditions, the wind farm configuration is represented as a graph in [106], and then the physics-induced GNNs are utilized to model the interaction among wind turbines.

IV. KEY ISSUES AND FUTURE DEVELOPMENT

Although GNNs have received extensive attention, and rich literature studies about this area have been published, developments in GNNs and the power system itself will certainly lead to new opportunities and problems. In this section, the main issues and future development are highlighted.

*A. Main issues*

The power systems are very complicated and have many uncertain factors. The successful applications of GNNs in computer visions prove that they can explore complex objective laws of high dimensional data through unsupervised learning. However, the existing parameters and structures of GNNs are designed for the data of computer vision, which is not suitable for the 1-dimensional time series in power systems. Therefore, it needs further research on how to adjust parameters and structures of GNNs with strong feature extraction ability and high-quality solutions according to the characteristics of the data from power systems.

Up to now, the application of GNNs in power systems has rarely been commercially viable and practical. On one side, the theory about GNNs is still not perfect and is still in the primary stage of exploration and verification. On the other side, the power systems have high requirements for stability and reliability of control approaches [107], while current GNNs are still based on statistical law and probability.

*B. Future development*

Traditional DNNs can get better performance by stacking hundreds of layers, since the deeper structure has more parameters, which significantly improves the representing ability. However, GNNs are suitable for shallow structures, and most of GNNs are no more than three layers. How to design real deep structures of GNNs is an exciting challenge for future research.

Another problem is how to deal with graph-structured data with dynamic structures. Static graph-structured data is stable, which can be modeled feasibly, while dynamic graph-structured data include changing structures. For example, the number of branches and nodes in power systems may often increase, and GNNs can't adapt to these changes. Therefore, finding the dynamic GNNs will be a big milestone about the adaptability and stability of general GNNs.

V. CONCLUSIONS

This paper provides a comprehensive review of GNNs in power systems, including several classical paradigms and their applications in fault scenario application, time series prediction, power flow calculation, and data generation. The main issues and future development in this area have been summarized and discussed. In general, GNNs and their application in power systems still face many challenges and opportunities, which will attract more attention and research. There will inevitably be more outstanding developments in the future.

REFERENCES

[1] B. Mohandes, M. Moursi, N. Hatziargyriou *et al*., "A review of power system flexibility with high penetration of renewables," *IEEE Transactions on Power Systems*, vol. 34, no. 4, pp. 3140-3155, Jul. 2019.
[2] J. Su, T. Lie, and R. Zamora, "Integration of electric vehicles in distribution network considering dynamic power imbalance issue," *IEEE Transactions on Industry Applications*, vol. 56, no. 5, pp. 5913-5923, Sept. 2020.
[3] Y. Chen, Y. Wang, D. Kirschen *et al*., "Model-free renewable scenario generation using generative adversarial networks," *IEEE Transactions on Power Systems*, vol. 33, no. 3, pp. 3265-3275, May. 2018.
[4] O. Alimi, K. Ouahada, and A. Abu-Mahfouz, "A review of machine learning approaches to power system security and stability," *IEEE Access*, vol. 8, pp. 113512-113531, Jun. 2020.
[5] X. Tao, D. Zhang, Z. Wang *et al*., "Detection of power line insulator defects using aerial images analyzed with convolutional neural networks," *IEEE Transactions on Systems, Man, and Cybernetics: Systems*, vol. 50, no. 4, pp. 1486-1498, Apr. 2020.
[6] W. Liao, Y. Yun, Y. Wang *et al*., "Reactive power optimization of distribution network based on graph convolutional network," *Power System Technology*, vol. PP, no. 99, pp. 1-12, Nov. 2020.
[7] Z. Li, Y. Xing, J. Huang *et al*., "Large-scale online multi-view graph neural network and applications," *Future Generation Computer Systems*, vol. 116, pp. 145-155, Mar. 2021.
[8] Y. Li, Q. Wang, X. Wang *et al*., "Community enhanced graph convolutional networks," *Pattern Recognition Letters*, vol. 138, pp. 462-468, Oct. 2020.
[9] K. Rusek, J. Suarez-Varela, P. Almasan *et al*., "RouteNet: leveraging graph neural networks for network modeling and optimization in SDN," *IEEE Journal on Selected Areas in Communications*, vol. 38, no. 10, pp. 2260-2270, Oct. 2020.
[10] Z. Zhang, P. Cui, and W. Zhu, "Deep learning on graphs: a survey," *IEEE Transactions on Knowledge and Data Engineering*, vol. PP, no. 99, pp. 1-1, Mar. 2020.
[11] J. Zhou, G. Cui, Z. Zhang *et al*. (20 Dec 2018) Graph neural networks: a review of methods and applications. [Online]. Available: https://arxiv.org/abs/1812.08434
[12] Z. Wu, S. Pan, F. Chen *et al*., "A comprehensive survey on graph neural networks," *IEEE Transactions on Neural Networks and Learning Systems*, vol. 32, no. 1, pp. 4-24, Jan. 2021.
[13] M. Nickel, K. Murphy, V. Tresp *et al*., "A review of relational machine learning for knowledge graphs," *Proceedings of the IEEE*, vol. 104, no. 1, pp. 11-33, Jan. 2016.
[14] W. Cao, Z. Yan, Z. He *et al*., "A comprehensive survey on geometric deep learning," *IEEE Access*, vol. 8, pp. 35929-35949, Feb. 2020.
[15] Z. Zhang, D. Zhang, and R. Qiu, "Deep reinforcement learning for power system applications: An overview," *CSEE Journal of Power and Energy Systems*, vol. 6, no. 1, pp. 213-225, Mar. 2020.
[16] D. Zhang, X. Han, and C. Deng, "Review on the research and practice of deep learning and reinforcement learning in smart grids," *CSEE Journal of Power and Energy Systems*, vol. 4, no. 3, pp. 362-370, Sept. 2018.
[17] M. Khodayar, G. Liu, J. Wang *et al*., "Deep learning in power systems research: a review," *CSEE Journal of Power and Energy Systems*, vol. PP, no. 99, pp. 1-13, Nov. 2020.
[18] M. Gori, G. Monfardini, and F. Scarselli, "A new model for learning in graph domains," in *Proceedings of 2005 IEEE International Joint Conference on Neural Networks*, Que., Canada, Dec. 2005, pp. 729-734.
[19] F. Scarselli, M. Gori, A. Tsoi *et al*., "The graph neural network model," *IEEE Transactions on Neural Networks*, vol. 20, no. 1, pp. 61-80, Jan. 2009.
[20] J. Bruna, W. Zaremba, A. Szlam et al., "Spectral networks and locally connected networks on graphs," in Proceedings of 2nd International Conference on Learning Representations, AB, Canada, Apr. 2014, pp. 1-14.
[21] A. Micheli, "Neural network for graphs: a contextual constructive approach," *IEEE Transactions on Neural Networks*, vol. 20, no. 3, pp. 498-511, Mar. 2009.
[22] V. N. Ekambaram, G. C. Fanti, B. Ayazifar *et al*., "Spline-like wavelet filterbanks for multiresolution analysis of graph-structured data," *IEEE Transactions on Signal and Information Processing over Networks*, vol. 1, no. 4, pp. 268-278, Dec. 2015.
[23] C. Tang, J. Sun, Y. Sun *et al.*, "A general traffic flow prediction approach based on spatial-temporal graph attention," *IEEE Access*, vol. 8, pp. 153731-153741, Aug. 2020.
[24] Y. Liu, Y. Liu and C. Yang, "Modulation recognition with graph convolutional network," *IEEE Wireless Communications Letters*, vol. 9, no. 5, pp. 624-627, May. 2020.
[25] M. Defferrard, X. Bresson, and P. Vandergheynst, "Convolutional neural networks on graphs with fast localized spectral filtering," in *Proceedings of the Neural Information Processing Systems 29 (NIPS 2016)*, Barcelona, Spain, Dec. 2016, pp. 1-9.
[26] R. Berg, T. Kipf, and M. Welling, "Graph convolutional matrix completion," in *Proceedings of KDD Deep Learning Day*, London, United Kingdom, Aug. 2018, pp. 1-7.
[27] T. Kipf and M. Welling, "Semi-supervised classification with graph convolutional networks," in *Proceedings of 5th International Conference on Learning Representations*, Toulon, France, Apr. 2017, pp. 1-14.
[28] A. Qin, Z. Shang, J. Tian *et al*., "Spectral–spatial graph convolutional networks for semisupervised hyperspectral image classification," *IEEE Geoscience and Remote Sensing Letters*, vol. 16, no. 2, pp. 241-245, Feb. 2019.
[29] Y. Yang and Y. Qi, "Image super-resolution via channel attention and spatial graph convolutional network," *Pattern Recognition*, vol. 112, pp. 1-1, Apr. 2021.
[30] D. Duvenaud, D. Maclaurin, J. Iparraguirre *et al*., "Convolutional networks on graphs for learning molecular fingerprints," in *Proceedings of the Neural Information Processing Systems 28*, Montreal, Canada, Dec. 2015, pp. 1-9.
[31] J. Atwood and D. Towsley, "Diffusion-convolutional neural networks," in *Proceedings of the Neural Information Processing Systems 29*, Barcelona, Spain, Dec. 2016, pp. 1-9.
[32] C. Zhuang and Q. Ma, "Dual graph convolutional networks for graph-based semi-supervised classification," in *Proceedings of 2018 World Wide Web Conference*, Lyon, France, Apr. 2018, pp. 499-508.

[33] M. Balcilar, G. Renton, P. Heroux et al. (2020, Mar.). Bridging the gap between spectral and spatial domains in graph neural networks. [Online]. Available: https://arxiv.org/abs/2003.11702

[34] R. Levie, F. Monti, X. Bresson et al., "CayleyNets: graph convolutional neural networks with complex rational spectral filters," *IEEE Transactions on Signal Processing*, vol. 67, no. 1, pp. 97-109, Jan. 2019.

[35] X. Wang, H. Ji, C. Shi et al., "Heterogeneous Graph Attention Network," in *Proceedings of World Wide Web Conference*, CA, USA, May. 2019, pp. 2022–2032.

[36] J. Gilmer, S. Schoenholz, P. Riley et al., "Neural message passing for quantum chemistry," in *Proceedings of the 34th International Conference on Machine Learning*, Sydney, Australia, Aug. 2017, pp. 1263-1272.

[37] T. Derr, Y. Ma, and J. Tang, "Signed graph convolutional networks," in *Proceedings of IEEE International Conference on Data Mining*, Sentosa island, Singapore, Nov. 2018, pp. 929-934.

[38] G. Melo, D. Sugimoto, P. Tasinaffo et al., "A new approach to river flow forecasting: LSTM and GRU multivariate models," *IEEE Latin America Transactions*, vol. 17, no. 12, pp. 1978-1986, Dec. 2019.

[39] Y. Li, D. Tarlow, M. Brockschmidt et al., "Gated graph sequence neural networks," in *Proceedings of the 4th International Conference on Learning Representations*, Puerto Rico, USA, May. 2016, pp. 1-20.

[40] K. Tai, R. Socher, and C. Manning, "Improved semantic representations from tree-structured long short-term memory networks," in *Proceedings of Association for Computational Linguistics*, Beijing, China, Jul. 2015, pp. 1-11.

[41] V. Ioannidis, A. Marques, and G. Giannakis, "Tensor graph convolutional networks for multi-relational and robust learning," *IEEE Transactions on Signal Processing*, vol. 68, pp. 6535-6546, Oct. 2020.

[42] P. Velickovic, G. Cucurull, A. Casanova et al., "Graph attention networks," in *Proceedings of the 6th International Conference on Learning Representations*, Vancouver, Canada, Apr. 2018, pp. 1-12.

[43] A. Vaswani, N. Shazeer, N. Parmar et al., "Attention is all you need," in *Proceedings of the Neural Information Processing Systems 30 (NIPS 2017)*, California, USA, Dec. 2017, pp. 1-15.

[44] J. Zhang, X. Shi, J. Xie et al., "GaAN: gated attention networks for learning on large and spatiotemporal graphs," in *the Conference on Uncertainty in Artificial Intelligence*, California, USA, Aug. 2018, pp. 1-10.

[45] W. Liu, P. Chen, F. Yu et al., "Learning graph topological features via GAN," *IEEE Access*, vol. 7, pp. 21834-21843, Feb. 2019.

[46] J. Feng, and S. Chen, "Link prediction based on orbit counting and graph auto-encoder," *IEEE Access*, vol. 8, pp. 226773-226783, Dec. 2020.

[47] Y. Ding, L. Tian, X. Lei et al., "Variational graph auto-encoders for miRNA-disease association prediction," *Methods*, vol. PP, no. 99, pp. 1-10, Aug. 2020.

[48] D. Valsesia, G. Fracastoro, and E. Magli, "Learning localized representations of point clouds with graph-convolutional generative adversarial networks," *IEEE Transactions on Multimedia*, vol. 23, pp. 402-414, Feb. 2020.

[49] Z. Wang, W. Wang, C. Liu et al., "Probabilistic forecast for multiple wind farms based on regular vine copulas," *IEEE Transactions on Power Systems*, vol. 33, no. 1, pp. 578-589, Jan. 2018.

[50] Y. Li, R. Yu, C. Shahabi et al., "Diffusion convolutional recurrent neural network: data-driven traffic forecasting," in *Proceedings of the 6th International Conference on Learning Representations*, Vancouver, Canada, Apr. 2018, pp. 1-16.

[51] Y. Seo, M. Defferrard, P. Vandergheynst et al., "Structured sequence modeling with graph convolutional recurrent networks," in *Proceedings of the Neural Information Processing*, Siem Reap, Cambodia, Nov. 2018, pp. 362-373.

[52] Y. Bi, A. Chadha, A. Abbas et al., "Graph-based spatio-temporal feature learning for neuromorphic vision sensing," *IEEE Transactions on Image Processing*, vol. 29, pp. 9084-9098, Sept. 2020.

[53] A. Jain, A. R. Zamir, S. Savarese et al., "Structural-RNN: deep learning on spatio-temporal graphs," in *Proceedings of the 2016 IEEE Conference on Computer Vision and Pattern Recognition*, NV, USA, Jun. 2016, pp. 5308-5317.

[54] Y. Zhang, Y. Li, X. Wei et al., "Adaptive spatio-temporal graph convolutional neural network for remaining useful life estimation," in *Proceedings of the 2020 International Joint Conference on Neural Networks*, Glasgow, United Kingdom, Jul. 2020, pp. 1-7.

[55] Sijie Yan, Yuanjun Xiong, and Dahua Lin, "Spatial temporal graph convolutional networks for skeleton-based action recognition," in *Proceedings of the 2020 International Joint Conference on Neural Networks*, Louisiana, USA, Feb. 2018, pp. 1-10.

[56] B. Yu, H. Yin, and Z. Zhu, "Spatio-temporal graph convolutional networks: a deep learning framework for traffic forecasting," in *Proceedings of the 27th International Joint Conference on Artificial Intelligence*, Stockholm, Sweden, Jul. 2018, pp. 3634-3640.

[57] P. Quang, Y. Aoul, and A. Outtagarts, "A deep reinforcement learning approach for vnf forwarding graph embedding," *IEEE Transactions on Network and Service Management*, vol. 16, no. 4, pp. 1318-1331, Dec. 2019.

[58] R. Miao, X. Zhang, H. Yan et al., "A dynamic financial knowledge graph based on reinforcement learning and transfer learning," in *Proceedings of the 2019 IEEE International Conference on Big Data*, CA, USA, Dec. 2019, pp. 5370-5378.

[59] K. Do, T. Tran, and S. Venkatesh, "Graph transformation policy network for chemical reaction prediction," in *Proceedings of the The 25th ACM SIGKDD Conference on Knowledge Discovery and Data Mining*, NY, USA, Aug. 2019, pp. 1-21.

[60] J. You, B. Liu, R. Ying et al., "Graph convolutional policy network for goal-directed molecular graph generation," in *Proceedings of the Neural Information Processing Systems*, Montreal, Canada, Dec. 2018, pp. 1-12.

[61] N. Cao, and T. Kipf, "MolGAN: an implicit generative model for small molecular graphs," in *Proceedings of the International Conference on Machine Learning*, Stockholm, Sweden, Jul. 2018, pp. 1-11.

[62] S. J. Pan, and Q. Yang, "A survey on transfer learning," *IEEE Transactions on Knowledge and Data Engineering*, vol. 22, no. 10, pp. 1345-1359, Otc. 2010.

[63] Q. Zhu, Y. Xu, H. Wang et al. (11 Sept 2020) Transfer learning of graph neural networks with ego-graph information maximization. [Online]. Available: https://arxiv.org/abs/2009.05204

[64] J. Lee, H. Kim, J. Lee et al., "Transfer learning for deep learning on graph-structured data," in *Proceedings of the Thirty-First AAAI Conference on Artificial Intelligence*, California, USA, Feb. 2017, pp. 2154–2160.

[65] Y. Benmohamed, M. Teguar and A. Boubakeur, "Application of SVM and KNN to duval pentagon 1 for transformer oil diagnosis," *IEEE Transactions on Dielectrics and Electrical Insulation*, vol. 24, no. 6, pp. 3443-3451, Dec. 2017.

[66] T. Wang, Y. He, B. Li et al., "Transformer fault diagnosis using self-powered RFID sensor and deep learning approach," *IEEE Sensors Journal*, vol. 18, no. 15, pp. 6399-6411, Aug. 2018.

[67] W. Liao, D. Yang, Y. Wang et al., "Fault diagnosis of power transformers using graph convolutional network," *CSEE Journal of Power and Energy Systems*, vol. 7, no. 2, pp. 241-249, Mar. 2021.

[68] N. Sapountzoglou, J. Lago, B. Schutter et al., "A generalizable and sensor-independent deep learning method for fault detection and location in low-voltage distribution grids," *Applied Energy*, vol. 276, pp. 1-22, Oct. 2020.

[69] J. Freitas and F. Coelho, "Fault localization method for power distribution systems based on gated graph neural networks," *Electrical Engineering*, vol. PP, no. 99, pp. 1-8, Feb. 2021.

[70] K. Chen, J. Hu, Y. Zhang et al., "Fault location in power distribution systems via deep graph convolutional networks," *IEEE Journal on Selected Areas in Communications*, vol. 38, no. 1, pp. 119-131, Jan. 2020.

[71] R. Ferrari, T. Parisini, and M. Polycarpou, "Distributed fault detection and isolation of large-scale discrete-time nonlinear systems: an adaptive approximation approach," *IEEE Transactions on Automatic Control*, vol. 57, no. 2, pp. 275-290, Feb. 2012.

[72] H. Khorasgani, A. Hasanzadeh, A. Farahat et al., "Fault detection and isolation in industrial networks using graph convolutional neural networks," in *Proceedings of the IEEE International Conference on Prognostics and Health Management*, CA, USA, Jun. 2019, pp. 1-7.

[73] E. Kabir, S. Guikema, and S. Quiring, "Predicting thunderstorm-induced power outages to support utility restoration," *IEEE Transactions on Power Systems*, vol. 34, no. 6, pp. 4370-4381, Nov. 2019.

[74] D. Owerko, F. Gama, and A. Ribeiro, "Predicting power outages using graph neural networks," in *Proceedings of IEEE Global Conference on Signal and Information Processing*, CA, USA, Nov. 2018, pp. 743-747.

[75] B. Karmakar and A. Pradhan, "Detection and classification of faults in solar pv array using thevenin equivalent resistance," *IEEE Journal of Photovoltaics*, vol. 10, no. 2, pp. 644-654, Mar. 2020.

[76] J. Fan, S. Rao, G. Muniraju et al., "Fault classification in photovoltaic arrays using graph signal processing," in *Proceedings of the IEEE Conference on Industrial Cyberphysical Systems*, Tampere, Finland, Jun. 2020, pp. 315-319.

[77] A. Mamun, M. Sohel, N. Mohammad *et al*., "A comprehensive review of the load forecasting techniques using single and hybrid predictive models," *IEEE Access*, vol. 8, pp. 134911-134939, Jul. 2020.

[78] A. Karimi, Y. Wu, M. Koyuturk *et al*., "Spatiotemporal graph neural network for performance prediction of photovoltaic power systems," in *Proceedings of the* Third Annual Conference on Innovative Applications of Artificial Intelligence, Vancouver, Canada, Feb. 2021, pp. 1-8.

[79] M. Khodayar, G. Liu, J. Wang *et al*., "Spatiotemporal behind-the-meter load and PV power forecasting via deep graph dictionary learning," *IEEE Transactions on Neural Networks and Learning Systems*, vol. PP, no. 99, pp. 1-15, Dec. 2020.

[80] M. Khodayar, S. Mohammadi, M.Khodayar *et al*., "Convolutional graph autoencoder: a generative deep neural network for probabilistic spatio-temporal solar irradiance forecasting," *IEEE Transactions on Sustainable Energy*, vol. 11, no. 2, pp. 571-583, Apr. 2020.

[81] B. Kan, G. Liu, K. Mahdi *et al*., "Distributed photovoltaic generation prediction based on graph machine learning," *Distribution & Utilization*, vol. 36, no. 11, pp. 20-27, Nov. 2019.

[82] M. Yu, Z. Zhang, X. Li *et al*., "Superposition graph neural network for offshore wind power prediction," *Future Generation Computer Systems*, vol. 113, pp. 145-157, Dec. 2020.

[83] R. Chen, J. Liu, F. Wang *et al*., "Graph neural network-based wind farm cluster speed prediction," in *Proceedings of the* IEEE 3rd Student Conference on Electrical Machines and Systems, Jinan, China, Dec. 2020, pp. 1-6.

[84] M. Khodayar, and J. Wang, "Spatio-temporal graph deep neural network for short-term wind speed forecasting," *IEEE Transactions on Sustainable Energy*, vol. 10, no. 2, pp. 670-681, Apr. 2019.

[85] C. Dinesh, S. Makonin, and I. Bajic, "Residential power forecasting using load identification and graph spectral clustering," *IEEE Transactions on Circuits and Systems II: Express Briefs*, vol. 66, no. 11, pp. 1900-1904, Nov. 2019.

[86] V. Bolz, J. Rueß, and A. Zell, "Power flow approximation based on graph convolutional networks," in *Proceedings of 18th IEEE International Conference On Machine Learning And Applications*, FL, USA, Dec. 2019, pp. 1679-1686.

[87] Q. Yang, A. Sadeghi, G. Wang *et al*., "Power system state estimation using gauss-newton unrolled neural networks with trainable priors," in *Proceedings of the* IEEE International Conference on Communications, Control, and Computing Technologies for Smart Grids, AZ, USA, Nov. 2020, pp. 1-6.

[88] D. Wang, K. Zheng, Q. Chen *et al*., "Probabilistic power flow solution with graph convolutional network," in *Proceedings of the* IEEE PES Innovative Smart Grid Technologies Europe, The Hague, Netherlands, Oct. 2020, pp. 1-5.

[89] N. Retière, D. Ha, and J. Caputo, "Spectral graph analysis of the geometry of power flows in transmission networks," *IEEE Systems Journal*, vol. 14, no. 2, pp. 2736-2747, Jun. 2020.

[90] B. Donon, R. Clement, B. Donnot *et al*., "Neural networks for power flow: Graph neural solver," *Electric Power Systems Research*, vol. 189, pp. 1-9, Dec. 2020.

[91] D. Owerko, F. Gama, and A. Ribeiro, "Optimal power flow using graph neural networks," in *Proceedings of the IEEE International Conference on Acoustics, Speech and Signal Processing*, Barcelona, Spain, Dec. 2020, pp. 5930-5934.

[92] C. Kim, K. Kim, P. Balaprakash *et al*., "Graph convolutional neural networks for optimal load shedding under line contingency," in *Proceedings of the IEEE Power & Energy Society General Meeting*, GA, USA, Aug. 2019, pp. 1-5.

[93] M. Khodayar, S. Mohammadi, M. Khodayar *et al*., "Convolutional graph autoencoder: a generative deep neural network for probabilistic spatio-temporal solar irradiance forecasting," *IEEE Transactions on Sustainable Energy*, vol. 11, no. 2, pp. 571-583, Apr. 2020.

[94] T. Wu, Y. Zhang, Y. Liu *et al*., "Missing data recovery in large power systems using network embedding," *IEEE Transactions on Smart Grid*, vol. 12, no. 1, pp. 680-691, Jan. 2021.

[95] M. Liang, Y. Meng, J. Wang *et al*., "FeederGAN: Synthetic feeder generation via deep graph adversarial nets," *IEEE Transactions on Smart Grid*, vol. PP, no.99, pp. 1-1, Sept. 2020.

[96] H. Yuan, J. Zhang, P. Xu *et al*., "Study on fast charging demand guidance in coupled power-transportation networks based on graph reinforcement learning," *Power System Technology*, vol. PP, no.99, pp. 1-9, Oct. 2020.

[97] P. Xu, Y. Pei, X. Zheng *et al*., "A simulation-constraint graph reinforcement learning method for line flow control," in *Proceedings of the* IEEE 4th Conference on Energy Internet and Energy System Integration, Wuhan, China, Oct. 2020, pp. 1-6.

[98] W. Miao1, H. Wu1, P. Chen *et al*., "Intelligent auxiliary operation and maintenance system of power communication network based on knowledge graph," in *Proceedings of the* 2020 *International Seminar on Artificial Intelligence*, Shanghai, China, Sept. 2020, pp. 1-11.

[99] F. Fusco, B. Eck, R. Gormally *et al*., "Knowledge and data-driven services for energy systems using graph neural networks," in *Proceedings of the* IEEE International Conference on Big Data, GA, USA, Dec. 2020, pp. 1-8.

[100] Kumthekar and Yashodhan. (22 May 2020) Using ChebConv and B-Spline GNN models for solving unit commitment and economic dispatch in a day ahead energy trading market based on ERCOT nodal model. [Online]. Available: https://rc.library.uta.edu/uta-ir/handle/10106/29117

[101] C. Wang, J. An, and G. Mu, "Power system network topology identification based on knowledge graph and graph neural network," *Fronts in Energy Research*, vol. 8, pp. 1-12, Feb. 2021.

[102] S. Füllhase, J.o Heres, and Y. Shapovalova. (1 Jul 2020) Testing the n-1 principle with graph neural networks. [Online]. Available: https://doi.org/10.13140/RG.2.2.28360.34569

[103] Z. Wang, Y. Chen, S. Huang *et al*., "Temporal Graph Super Resolution on Power Distribution Network Measurements," *IEEE Access*, vol. PP, pp. 1-11, Jan. 2021.

[104] J. Yu, D. Hill, V. Li *et al*., "Synchrophasor recovery and prediction: a graph-based deep learning approach," *IEEE Internet of Things Journal*, vol. 6, no. 5, pp. 7348-7359, Oct. 2019.

[105] J. Huang, L. Guan, Y. Su *et al*., "Recurrent graph convolutional network-based multi-task transient stability assessment framework in power system," *IEEE Access*, vol. 4, pp. 93283-93296, Apr. 2020.

[106] J. Park and J. Park, "Physics-induced graph neural network: An application to wind-farm power estimation," *Energy*, vol. 187, pp. 1-15, Nov. 2019.

[107] L. Duchesne, E. Karangelos, and L. Wehenkel, "Recent developments in machine learning for energy systems reliability management," *Proceedings of the IEEE*, vol. 108, no. 9, pp. 1656-1676, Sept. 2020.

**Wenlong Liao** received the B.S. degree from China Agricultural University, Beijing, China, in 2017. He received the M.S. degree from Tianjin University, Tianjin, China, in 2020. He is currently pursuing the Ph.D. degree in Aalborg University, Aalborg, Denmark. His current research interests include smart grids, machine learning, and renewable energy.

**Birgitte Bak-Jensen** received the M.Sc. degree in electrical engineering and the Ph.D. degree in modeling of high voltage components from the Institute of Energy Technology, Aalborg University, Aalborg, Denmark, in 1986 and 1992, respectively. She is currently a Professor of intelligent control of the power distribution system, Department of Energy Technology, Aalborg University. Her current research interests include power quality and stability in power systems and taking integration of dispersed generation and smart grid issues like demand response into account. Also, the interaction between the electrical grid and the heating and transport sector is a key area of interest.

**Jayakrishnan Radhakrishna Pillai** received the M.Tech. degree in power systems from the National Institute of Technology, Calicut, India, in 2005, the M.Sc. degree in sustainable energy systems from the University of Edinburgh, Edinburgh, U.K., in 2007, and the Ph.D. degree in power systems from Aalborg University, Aalborg, Denmark, in 2011. He is currently an Associate Professor with the Department of Energy Technology, Aalborg University. His current research interests include distribution system analysis, grid integration of electric vehicles and distributed energy resources, smart grids, and intelligent energy systems.

**Yuelong Wang** received the B.S. degree from Wuhan University, Wuhan, China, in 2017. He received the M.S. degree from Tianjin University, Tianjin, China, in 2020. He is an engineer who currently works for State Grid Tianjin Chengxi Electric Power Supply Branch, Tianjin, China. His current research interests include smart grids renewable energy, and machine learning.

**Yusen Wang** received the B.S. degree from China Agricultural University, Beijing, China, in 2017. He received the M.S. degree from KTH Royal Institute of Technology, Stockholm, Sweden, in 2019. He is currently pursuing the Ph.D. degree in KTH Royal Institute of Technology, Stockholm, Sweden. His current research interests include machine learning, and fault diagnosis for transformer and power quality.